\documentclass[conference]{IEEEtran}

\usepackage{graphicx}
\usepackage{epstopdf}
\usepackage{subcaption}
\usepackage{dblfloatfix}
\usepackage{color}

\title{Forward Table-Based Presynaptic Event-Triggered Spike-Timing-Dependent Plasticity}

\author{\IEEEauthorblockN{Bruno U. Pedroni\IEEEauthorrefmark{1},
Sadique Sheik\IEEEauthorrefmark{1},
Siddharth Joshi\IEEEauthorrefmark{1},
Georgios Detorakis\IEEEauthorrefmark{2},\\
Somnath Paul\IEEEauthorrefmark{3},
Charles Augustine\IEEEauthorrefmark{3},
Emre Neftci\IEEEauthorrefmark{2},
and Gert Cauwenberghs\IEEEauthorrefmark{1}}

\IEEEauthorblockA{\IEEEauthorrefmark{1}University of California, San Diego\\La Jolla, CA, USA 92093, Email: bpedroni@eng.ucsd.edu, gert@ucsd.edu}

\IEEEauthorblockA{\IEEEauthorrefmark{2}University of California, Irvine\\Irvine, CA, USA 92697, Email: eneftci@uci.edu}

\IEEEauthorblockA{\IEEEauthorrefmark{3}Intel Corporation - Circuit Research Lab\\ Hillsboro, OR, USA 97124, Email: somnath.paul@intel.com}}

\begin{document}

\maketitle

\begin{abstract}
Spike-timing-dependent plasticity (STDP) incurs both causal and acausal synaptic weight updates, for negative and positive time differences between pre-synaptic and post-synaptic spike events. For realizing such updates in neuromorphic hardware, current implementations either require forward and reverse lookup access to the synaptic connectivity table, or rely on memory-intensive architectures such as crossbar arrays.
We present a novel method for realizing both causal and acausal weight updates using only forward lookup access of the synaptic connectivity table, permitting memory-efficient implementation.  A simplified implementation in FPGA, using a single timer variable for each neuron, closely approximates exact STDP cumulative weight updates for neuron refractory periods greater than 10 ms, and reduces to exact STDP for refractory periods greater than the STDP time window. Compared to conventional crossbar implementation, the forward table-based implementation leads to substantial memory savings for sparsely connected networks supporting scalable neuromorphic systems with fully reconfigurable synaptic connectivity and plasticity.
\end{abstract}

\section{Introduction}

Neuromorphic systems are electronic instantiations of biological nervous systems which seek to mimic behavioral and structural aspects of real neural networks \cite{indiveri2011neuromorphic}. The three main components of neuromorphic systems are: neurons (representing processors), synapses (representing memory), and a learning rule. Neuromorphic systems differentiate themselves from traditional von Neumann architectures mainly due to distributed memory and parallel processing. Distributing the entire neural network of a neuromorphic chip into many smaller networks (i.e. \emph{cores}) permits more efficient spike event routing and memory access \cite{merolla2014million}.

\begin{figure}[!h]
  \centering
  \includegraphics[width=0.49\textwidth]{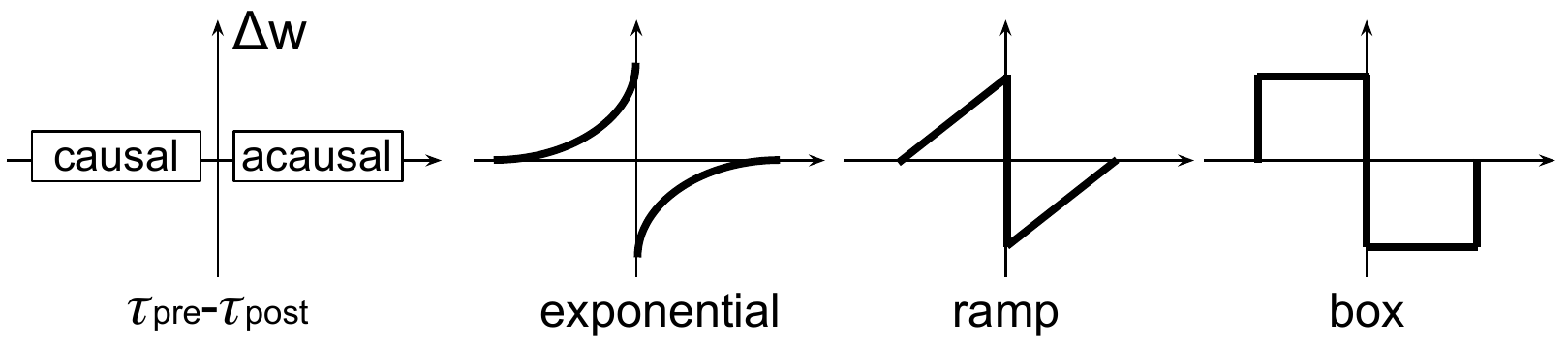}
  \caption{The causal and acausal weight update regions of the STDP kernels (left) and three typical kernels implemented in digital neuromorphic systems.}
  \label{fig:kernels}
  \vspace{-13pt}
\end{figure}

Learning in neuromorphic systems is usually realized by adapting the synaptic strength (or \emph{weight}) between neurons. Among observed forms of synaptic plasticity in biological nervous systems, spike-timing-dependent plasticity (STDP) \cite{bi1998synaptic,sjostrom2010spike} is particularly attractive for neuromorphic systems due to its locality property: weight adaptation requires only information of neighboring neurons. STDP relies on the relative temporal difference between pre-synaptic and post-synaptic spike events to adjust the direction and intensity of synaptic strength between neurons \cite{bi1998synaptic}. In the original STDP formulation, the intensity of weight updates depends on the temporal difference between spikes and on the current value of the weight. For simplicity, digital neuromorphic implementations of STDP usually consider a variant of the original STDP where the weight updates are simply a function of the temporal difference \cite{cassidy2011combinational}. Fig. \ref{fig:kernels} illustrates three typical STDP kernels implemented in digital neuromorphic systems. In the three cases, causal events (i.e. when pre- precedes post-synaptic spikes) produce weight increase, while acausal events (i.e. when post- precedes pre-synaptic spikes) produce weight decrease.

\vspace{-3pt}

\begin{figure*}[t]
  \centering
  \begin{subfigure}[b]{0.58\textwidth}
    \includegraphics[width=\textwidth]{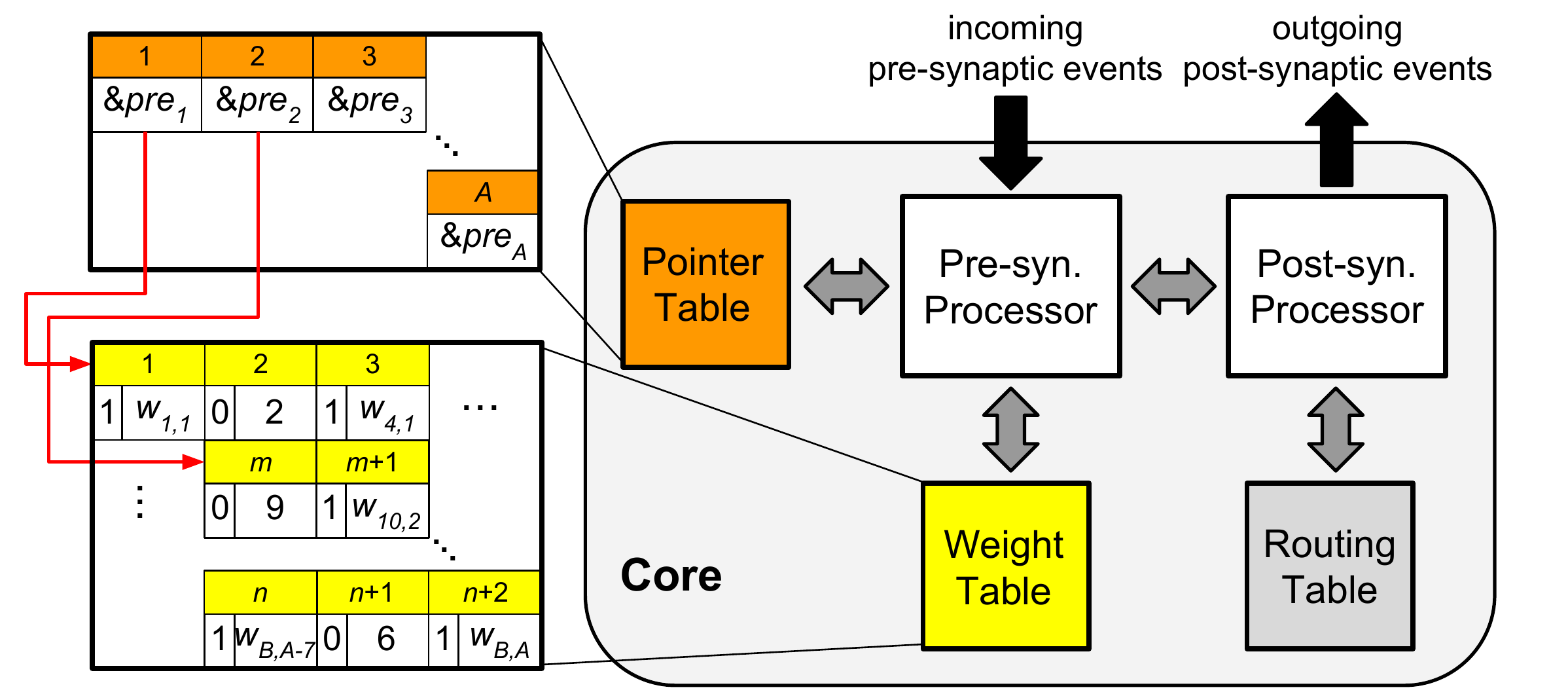}
    \caption{}
    \label{fig:index_core}
    \end{subfigure}%
    \hspace{+10pt}
    \begin{subfigure}[b]{0.35\textwidth}
    \includegraphics[width=\textwidth]{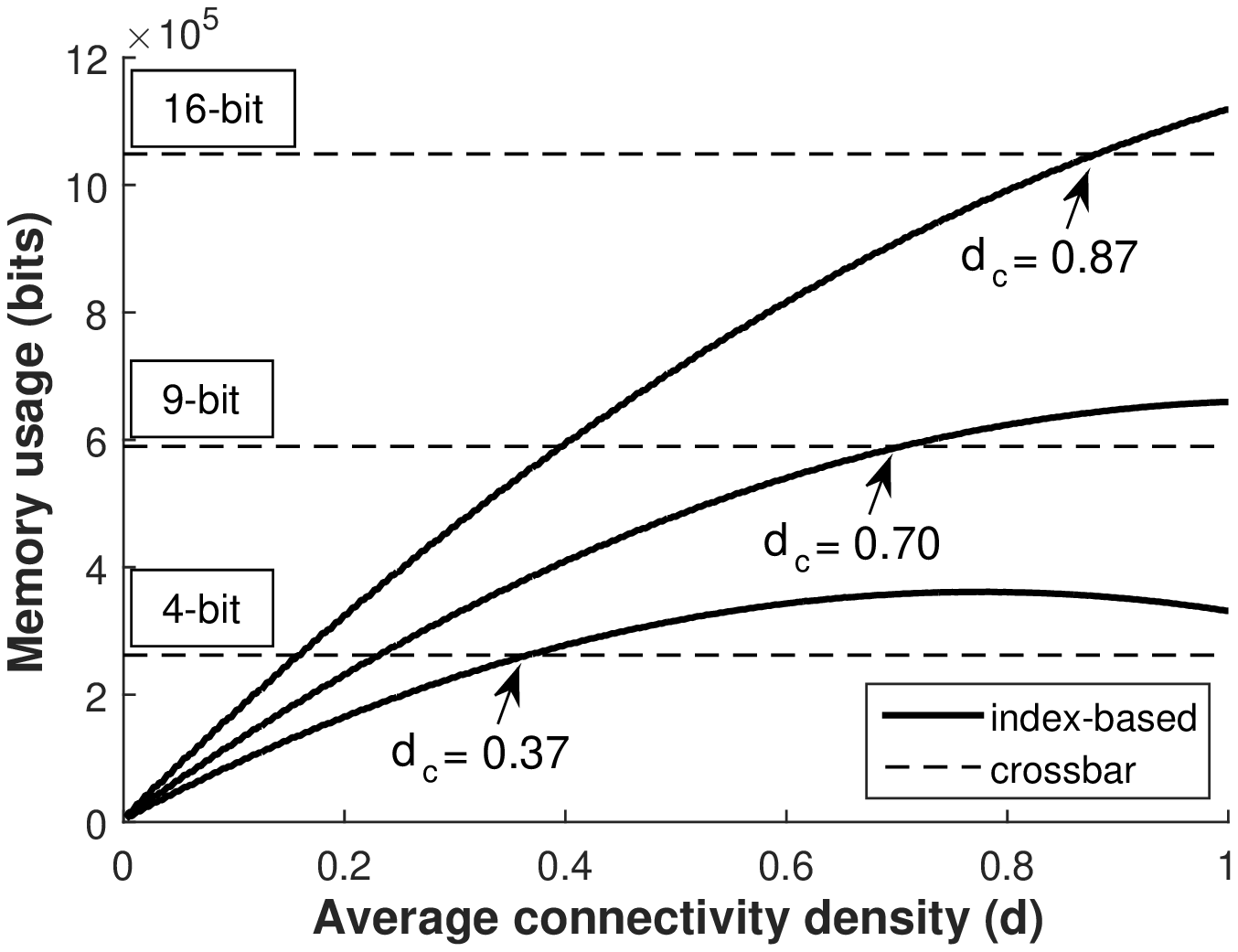}
    \caption{}
    \label{fig:comparison}
    \end{subfigure}
	\vspace{-9pt}
\caption{(a) The index-based architecture. Besides the weight table (WT) and routing table (RT), data compression is obtained by means of the pointer table (PT). In the example, the WT uses run-length encoding (RLE) with leading bit `0' indicating the run count (i.e. the number of consecutive post-synaptic neurons which the pre-synaptic neuron is not connected to), and leading bit `1' indicating an existing connection. (b) Memory usage comparison between the crossbar and the index-based core architectures based on the connectivity density between pre- and post-synaptic neurons.}
	\vspace{-18pt}
\end{figure*}

In biological systems, synaptic connectivity is instantiated at a physical level: there is a physical path between neurons. Though there have been multiple neuromorphic implementations of physically-connected neurons using, mainly, memristor \cite{jo2010nanoscale} and phase-change memory approaches \cite{kuzum2011nanoelectronic}, most of the large scale systems developed to date make use of (digital) random access memories representing ``virtual'' synaptic connections. These virtual connections are basically composed of two tables: routing and weight. The routing tables (RTs) are used to store the destination address of newly generated spikes, indicating to which core inputs they should be delivered. The weight tables (WTs) include pairs of synaptic connectivity and strength between input spikes (from other post-synaptic neurons) and their destination post-synaptic neurons in the core. Lastly, by organizing the WTs in specific manners inside each core, different neuromorphic architectures can be obtained. The most straightforward representation is by means of a crossbar architecture, in which all incoming pre-synaptic neurons are connected to all post-synaptic neurons in a core \cite{merolla2014million}. However, when we deal with sparse representations (i.e. less than 70\% connection density) or non-systematic connectivity patterns between neurons (i.e. each neuron connects to a different number of neurons inside the same core), we show that a more optimized solution is to use an index-based architecture. The specificities of this architecture are described in the following section. The remainder of the paper includes a detailed description of our proposed method, along with simulation and emulation results, and finally conclusions.

\section{Core architecture and processing}

A digital neuromorphic core can be interpreted as a neurosynaptic processor which includes an input-output weight map between pre- and post-synaptic neurons. The spikes produced by post-synaptic neurons are delivered to their destinations (on the same core or another core in the system) according to the core's routing table (RT). At the destination core, an incoming spike is received by one of its inputs, which in turn is connected to the post-synaptic neurons in the core via the weight table (WT). Therefore, each core input line can be interpreted as an axon, capable of fanning-out to multiple post-synaptic neurons via synaptic connections \cite{merolla2014million}. Digital neuromorphic systems normally operate in discrete time steps, or ``ticks''. These ticks are usually in the order of a millisecond, which is similar to the processing timescale of biological neurons. At the occurrence of a tick, the cores process only the inputs which have received a spike (i.e. event-based processing). Analogously, inputs which have not received spikes are skipped, reducing processing time and power consumption.


\subsection{Index-based core architecture}

The organization of the memory in the WT defines the type of core architecture. The two basic architectures are: crossbar and index-based. In the crossbar architecture every connection between an input and a post-synaptic neuron has a reserved space in the memory, even if the connection is not used in a given neural network. For very dense networks, the crossbar is the ideal solution. On the other hand, for sparser networks, where some connections between neurons are non-existent (i.e. they are of weight zero and will remain this way after STDP learning), or for topologies requiring each neuron to have different fan-outs, the index-based architecture is a better solution. Data stored in index-based architectures is compressed such that only WT memory for connections that exist in the neural network is used. This flexibility arises, however, at the cost of an additional table, the pointer table (PT), which contains a pointer for each core input to indicate its starting position in the WT. Fig. \ref{fig:index_core} exemplifies an index-based core with $A$ inputs and $B$ post-synaptic neurons.

To further improve processing and memory demands, lossless compression of the connections in the WT can be realized by using run-length encoding (RLE), where sequences of consecutive non-existent connections can be stored as run counts -- instead of a zero-valued weight per connection. An additional bit before each existing weight to indicate if the connection between an input and a post-synaptic neuron exists. For non-existent connections, the opposite valued bit can precede the run-length to indicate how many post-synaptic neurons should be skipped. In Fig. \ref{fig:index_core}, the WT was constructed with RLE using a `0' to indicate how many post-synaptic neurons to skip, and a `1' to indicate a connection that actually exists. With this, using RLE compression has the additional advantage over the crossbar architecture in that it decreases the number of post-synaptic neurons which must be analyzed for every incoming spike, reducing once again processing time and power consumption.

\subsection{Core architectures comparison}

For comparison purposes between the crossbar and index-based approaches, Fig. \ref{fig:comparison} shows the plots of memory usage (in bits) based on the connectivity density, $d$, between inputs and post-synaptic neurons. Core parameters of 256 inputs $\times$ 256 neurons with 9-bit weights were chosen as reference based on the work in \cite{merolla2014million}; 4-bit and 16-bit weights were also analyzed. For networks with density greater than the critical density ($d_c$), the crossbar architecture consumes less memory to map the connections and weights; this is due mainly to the additional pointer table and to the RLE overhead in the index-based approach. However, for connectivity densities below the critical density (e.g., $d_c=0.70$ for 9-bit weights), the index-based architecture shows better results. Naturally, cores with high bit-precision weights favor the index-based architecture since compression of non-existent connections becomes even more meaningful. The ``left over'' memory obtained using the index-based approach can be used, for example, to increase the weight bit-precision further, to better represent weights in the network, or perhaps, if an entire sector of the memory will not be required, power gating might also be a possibility as a means of reducing power consumption.

In theory, a neuromorphic system could use either type of memory organization: crossbar or index-based. This simply requires that the compiler efficiently choose the best architecture based on the core parameters and network topology being mapped, along with hardware being able to realize both types of memory accesses. One important aspect is that the crossbar approach has the advantage of being able to realize a reverse lookup, which is vital for the original STDP algorithm. However, as Fig. \ref{fig:comparison} shows, many networks are more efficiently mapped using the index-based approach, where we have access only to the forward connectivity. Therefore, as we will see in the next section, our novel implementation of STDP can be realized using simply this forward mapping and still produce results nearly identical to those of the original algorithm.


\section{Spike-timing dependent plasticity without reverse connectivity table lookup}

A fundamental aspect of the crossbar and index-based architectures for STDP is that a core has information about the pre-synaptic neuron spike times: an event delivered to an input in the core carries the temporal information about the pre-synaptic neuron, independently of where it is located in the system. This feature is vital for the STDP algorithm since it requires knowledge of spike times of both the post- and pre-synaptic neurons. In the digital neuromorphic core, this is realized by including an STDP timer for each input and post-synaptic neuron, representing the STDP learning window. Pre-synaptic (i.e. input) timers are initialized at the arrival of a spike, while post-synaptic timers are initialized at the generation of a new spike by post-synaptic neurons. If the causal and acausal STDP windows are not symmetric, then the STDP timer is set to the longest of the two windows. At each system tick, the STDP timer is decremented until it reaches zero (or a value greater than zero if the specific window is the shorter of the two). This implementation using a single timer for each neuron realizes the nearest-neighbor STDP learning rule since only the most recent spike events are considered.

In the traditional STDP algorithm, whenever a post-synaptic neuron fires, the causal weight updates are immediately executed. This is only possible, however, by knowing all the pre-synaptic neurons the specific post-synaptic neuron is connected to (i.e., by being able to perform reverse lookup access to the connectivity table) \cite{vogelstein2002spike}. Since the index-based architecture has access only to forward connectivity, in our method the causal update is delayed until the pre-synaptic STDP timer expires. With this, all weight updates are performed at the start or the end events of the pre-synaptic timer. Previous work \cite{jin2010implementing} demonstrated forward table-based STDP by deferring both causal and acausal updates, making use of system delays in spike time storage allowed by constraints on firing rates of pre- and post-synaptic neurons. The proposed method here performs the acausal weight updates at the onset of a pre-synaptic spike. This results in simplified spike time storage and reduces memory requirements by using the same STDP window for both causal and acausal updates.

\begin{figure}[!t]
	\centering
	\includegraphics[width=0.45\textwidth]{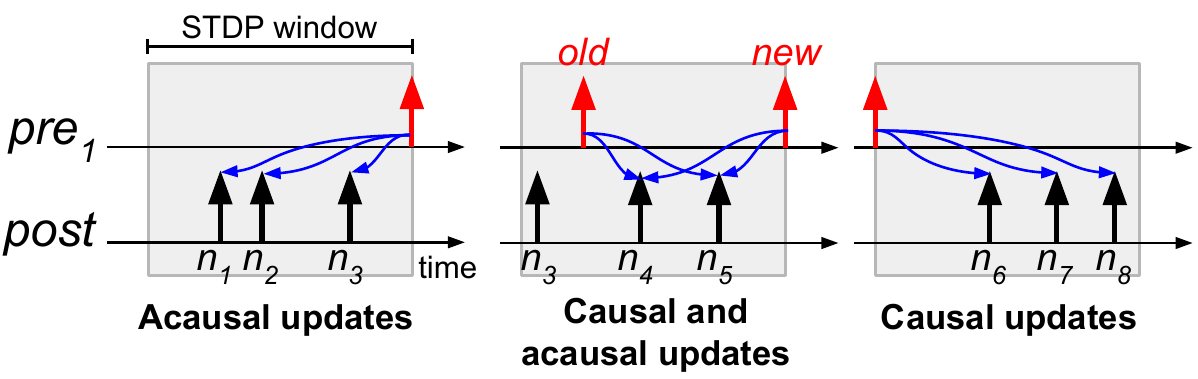}
	\caption{The proposed STDP weight update rule. (left) New pre-synaptic input events produce acausal updates. (center) Causal updates must be processed just before acausal updates when a previous event is still present. (right) Ending events produce delayed causal updates.}
	\label{fig:stdp}
    \vspace{-21pt}
\end{figure}

Figure \ref{fig:stdp} illustrates the moments when weight updates are performed using our method; for clarity, we used causal and acausal windows of same duration. In the figure, the spike event queue of pre-synaptic input 1 is represented in the top row ($pre_1$), while the spike event queue of all the post-synaptic neurons in this same core is represented in the bottom row ($post$). The indices below the bottom row spikes indicate the post-synaptic neuron address. The algorithm for our STDP learning rule is summarized by the following cases:

\vspace{-3pt}
\begin{enumerate}
 \item When a new pre-synaptic input spike arrives, perform the acausal weight updates;

  \item If the pre-synaptic STDP timer has not expired as this same  pre-synaptic input receives a new spike, perform the causal weight updates, followed by the acausal updates in case 1;
  
  \item When a pre-synaptic STDP timer expires, perform the causal weight updates.
\end{enumerate}
\vspace{-3pt}

\noindent As shown in Fig. \ref{fig:stdp} (left), upon a new event at pre-synaptic input 1 (depicted by the red arrow in $pre_1$), we realize the acausal updates by sweeping through this input's entries in the WT using forward lookup and updating the weights between all its post-synaptic neurons ($post$) which present STDP timers greater than zero (depicted by black arrows). If, however, $pre_1$'s timer were greater than zero, as depicted in Fig. \ref{fig:stdp} (center), then we would first realize the causal updates between $pre_1$ and post-synaptic neurons who spiked after $pre_1^{old}$, followed by the acausal updates for $pre_1^{new}$ (as performed in Fig. \ref{fig:stdp} (left)). Lastly, the causal updates shown in Fig. \ref{fig:stdp} (right) occur in a similar fashion, except only when the pre-synaptic input STDP timer expires (i.e. when the red arrow leaves the STDP window). 

The proposed method delays the causal weight updates since they only occur at the end of pre-synaptic timers. Nonetheless, this effect on the convergence of the learning rule can be mediated by choosing small enough learning rates. The only other aspect in which our method differs from the original STDP algorithm is for a particular causal update: In the case when a pre-synaptic input event has already entered the STDP window, and was followed by a post-synaptic spike, no updates should occur between the input and this neuron until the pre-synaptic input timer expires. However, if the post-synaptic neuron spikes again while its timer is greater than zero, the ``old'' post-synaptic spike would be overwritten by the new one (by reinitializing the timer). With this, the causal update of the pre-synaptic input with the first (i.e. nearest-neighbor) post-synaptic spike would be lost.

A rather costly solution to realizing these nearest-neighbor causal updates would be to go through every pre-synaptic input whose timer is greater than zero yet smaller than the STDP timer of the post-synaptic neuron of interest, and going through each of these inputs' table to verify if they are connected to the specific post-synaptic neuron. The worst case time expenditure for this scenario is ``number of inputs'' $\times$ ``number of post-synaptic neurons''. A more objective solution is to ignore these updates altogether if we consider that a single pre-synaptic spike should not have a strong causal relation to a post-synaptic neuron which is firing very frequently. As we will later show in our results, this can indeed be considered. A final and exact solution would be to configure the neurons in the system with a refractory period of same or longer duration than that of the STDP time window. This would, therefore, impede a bursting behavior, and neurons would not be able to spike more than once during the STDP time window.


\section{Results}

\begin{figure}[!h]
	\centering
	\includegraphics[width=0.65\textwidth]{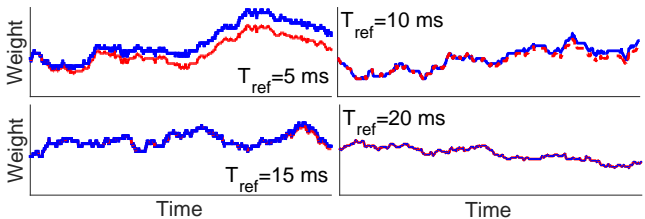}
    \vspace{-25pt}
	\caption{Evolution of training for the original STDP algorithm (blue) and our proposed method (red). As the refractory period approximates the STDP window duration of 20 ms, both methods converge.}
	\label{fig:results}
    \vspace{-12pt}
\end{figure}

For validation of our proposed method, we designed a $64$-input $\times$ $64$-neuron core, operating at 1-ms timesteps, on a Xilinx Spartan-6 FPGA. The emulation parameters were chosen to match that of biological scale. The anti-symmetric ramp STDP kernel depicted in Fig. \ref{fig:kernels}, with causal and acausal windows, $T_{stdp}$, of 20 ms each and max($\Delta w$) $ = \pm 1$, was used for our learning. The $4096$ 9-bit weights were initialized to value zero. Poisson spike trains with firing rate of 10 Hz and refractory periods, $T_{ref}$, varying between 5 and 20 ms were used to externally produce the pre- and post-synaptic spikes. The system was run for 60 seconds, after which we compared the weights obtained from the FPGA with simulation results of the original STDP algorithm. Fig. \ref{fig:results} compares the evolution of training with the original algorithm (in blue) and our proposed method (in red) for 4 different connections, with varying refractory periods. The figure shows how, as the refractory period approximates the duration of the causal and acausal windows, both algorithms converge.

\vspace{-5pt}

\begin{figure}[!h]
	\centering
	\includegraphics[width=0.49\textwidth]{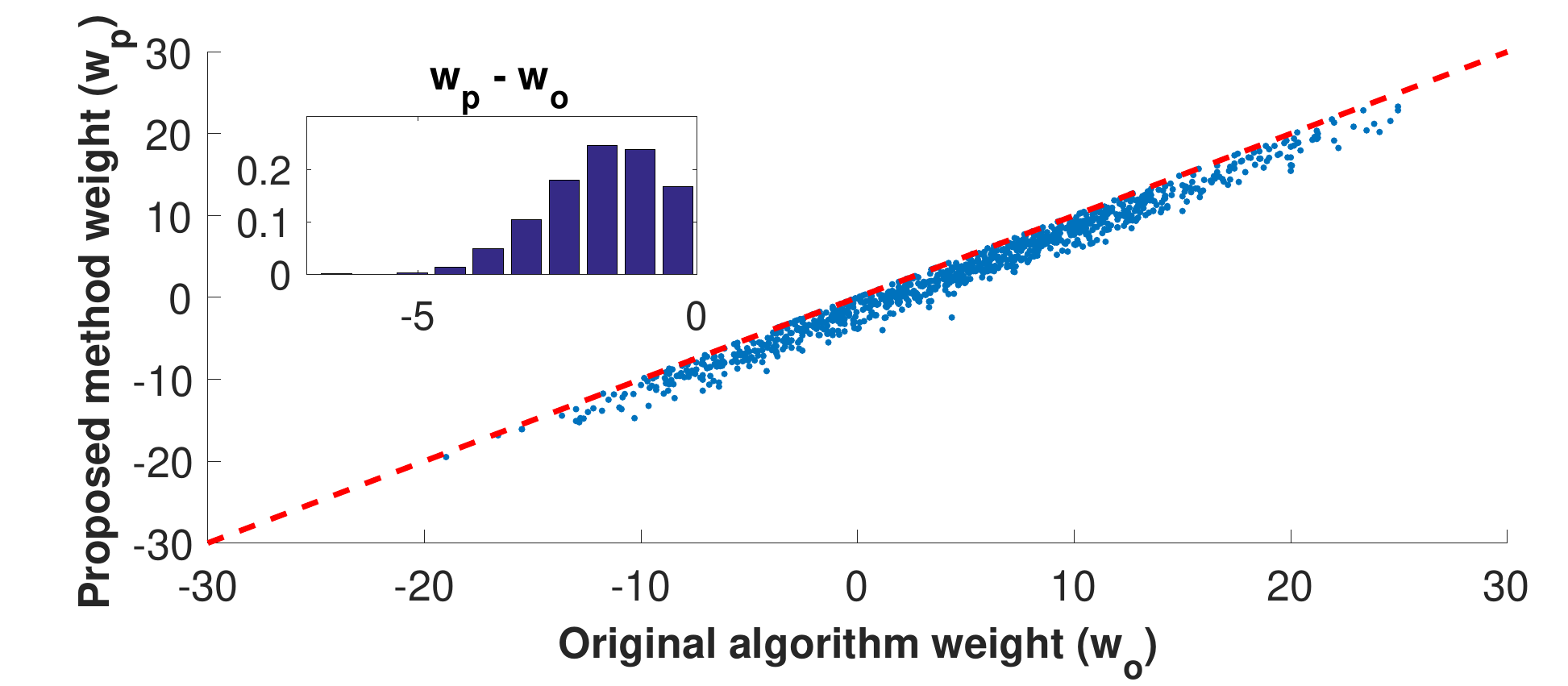}
	\caption{Comparison of final weights. Each blue point is a ($w_o$,$w_p$)-pair of weights produced by the original STDP algorithm and our proposed method, respectively.}
	\label{fig:errors}
    \vspace{-15pt}
\end{figure}

After running the system, using $T_{ref}=5$ ms, for 60 seconds, we also compared the final value of the weights produced by both algorithms. In Fig. \ref{fig:errors}, the blue points are pairs of weights produced by the original STDP algorithm ($w_o$) versus those produced by the FPGA after training ($w_p$); the red dashed line is the ideal scenario, where both methods match. The figure shows that our method produces slightly more negative weights. This is because in the FPGA implementation we are not performing the nearest-neighbor causal updates for post-synaptic neurons firing in rapid succession, resulting in a not so intense final update. However, as we see in the histogram in the figure, final weights with small difference between the ideal case and our method are much more frequent, and final weights with difference larger than $-4$ seldom occur.

\vspace{-7pt}

\section{Conclusions}

While the traditional spike-timing-dependent plasticity algorithm uses both forward and reverse connectivity tables, we presented a novel method of implementing STDP which benefits from event-driven operation in a very memory-efficient environment. Using simply forward lookup access to the connectivity table (halving the normal memory requirements for STDP implementation), causal weight updates can be performed at the moment of the pre-synaptic STDP timer expiration. Though this delayed update possesses an inherent caveat, we showed that, by ignoring certain weight updates because of the stronger influence of neighboring spikes, this issue can be mitigated. Additionally, if we are to configure neurons with refractory period equal to or longer than the STDP time window, then our method converges to the original algorithm.











\end{document}